\documentclass{article}
\usepackage{aaai}
\usepackage{times}
\usepackage{helvet}
\usepackage{courier}
\usepackage[sort, numbers]{natbib}
\usepackage{graphicx}

\begin{document}



\title{A Controllable Co-Creative Agent for Game System Design}
\author{Rohan Agarwal, Zhiyu Lin, Mark Riedl\\
Georgia Institute of Technology \\
\{roaga, zhiyulin\}@gatech.edu, riedl@cc.gatech.edu
}
\nocopyright 
\maketitle
\begin{abstract}
\begin{quote}
Many advancements have been made in procedural content generation for games, and with mixed-initiative co-creativity, have the potential for great benefits to human designers. However, co-creative systems for game generation are typically limited to specific genres, rules, or games, limiting the creativity of the designer. We seek to model games abstractly enough to apply to any genre, focusing on designing game systems and mechanics, and create a controllable, co-creative agent that can collaborate on these designs. We present a model of games using state-machine-like components and resource flows, a set of controllable metrics, a design evaluator simulating playthroughs with these metrics, and an evolutionary design balancer and generator. We find this system to be both able to express a wide range of games and able to be human-controllable for future co-creative applications.

\end{quote}
\end{abstract}

\section{Introduction}
Game content generation has made many recent advancements with more powerful generative models. Procedural content generation in some forms is already common in many popular games, and with current research, it is possible for both novices and professionals alike to create new levels of decent quality from scratch for various games \cite{pcgsurveyandgamedesign}, from Lode Runner \cite{bhaumik2021lode} to Super Mario Bros \cite{guzdial2016game}. Such tools range from completely automatic generation \cite{guzdial2016game} to collaborative systems involving a human designer \cite{bhaumik2021lode}.

Research in co-creation between a human designer and an AI system, including but not limited to game content creation, has led to promising results. A proper co-creative system with the right communications between human and AI can lead to superior creative product, novel ideas, and more human satisfaction, beyond what either human or AI could do alone \cite{Yannakakis2014MixedinitiativeC}. Co-creative systems, when matured, have great potential benefits for various creative industries, such as  improving creative works, decreasing time to produce a work, and increasing accessibility for designers. Some argue that AI can never replace human creativity, but no matter what, it is important for humans to be able to control AI and use it to get the creative works they want. For these reasons, human-AI co-creativity is an important area of research.

However, while a co-creative system for storytelling would usually let you write about anything \cite{lin2021plug}, one for game creation would usually let you create only more content for one kind of game--for example, an AI that generates Lode Runner levels \cite{bhaumik2021lode} or a 2D grid-based map designer \cite{Yannakakis2014MixedinitiativeC}. While it is undoubtedly useful to be able to create more content for a game, another major role of game designers is to design the main mechanics of a game in the first place. To have true benefit for game designers, the AI agent in such a system should be able to help humans design systems, mechanics, and gameplay loops for whatever creative intent they might have. Since these can be anything, the AI would be able to process "game design" \cite{pcgsurveyandgamedesign} as abstractly as possible.

We present a novel method for modeling game systems as state machines with resource flows, which we argue can be used to design mechanics of any kind for most kinds of games. Unlike existing models, we place no genre- or game-specific constraints, making it a method suitable for "game design," not "level design" or "content creation." We also present a human-controllable AI system using reinforcement learning and genetic algorithms that can manipulate and generate these designs, by balancing them, modifying pieces of them, generating new pieces, and providing insights to the human on the design. These abilities make our system suitable for co-creative applications.  We also introduce metrics for evaluating designs which are also used by the AI system. We analyze this system's capabilities to evaluate its potential.

\section{Background and Related Work}

\subsection{Co-Creative Systems}
Co-creativity between a human and an AI simply describes the two creating a single creative work together, whether it is a story \cite{lin2021plug}, game level \cite{bhaumik2021lode}, or drawing \cite{cocreativedrawing}. To truly elevate the creative experience in a beneficial way, mixed-initiative co-creative systems, where both human and AI explore, act, and contribute as collaborators, has been shown to be the most effective kind of co-creation \cite{Yannakakis2014MixedinitiativeC}. This work makes use of the mixed-initiative approach.

When studying or designing mixed-initiative systems, it is important to have an AI that is capable of making such contributions, but it is equally as important to consider the communications and interactions between the collaborators. There are several frameworks that define the possibility space and characteristics of such interactions. One is the Co-Creative Framework for Interaction design (COFI), which breaks down interaction into interaction with the product or between collaborators, and breaks each of those down even further into more sub-categories for understanding interaction \cite{COFI}. Another framework is Creative Wand, which presents an ontology of three dimensions of communication between the AI agent and human designer (elaborative/reflective, global/local, and human-/agent-initiaed), as well as a modular Python framework for building co-creative systems \cite{Lin_Agarwal_Riedl_2022}.

\subsection{Modeling Games}
One goal of this work is to be able to model game design in an abstract enough way for a co-creative system to be able to produce designs without genre, mechanical, or other constraints. Game design itself can be though of as a combination of system design (the rules, actions, and mechanics of a game) and world design \cite{pcgsurveyandgamedesign}. This paper focuses on system design.

There are several existing methods of modeling games, beyond specific models for specific games. One of the most popular methods is the Video Game Description Language (VGDL) \cite{VGDL}. VGDL can describe many grid-based 2D games with rules, events, a world layout, objects, and physics. It is also easily readable by both humans and machines. While this allows for some basic system design, it is a language more focused on creating specific game content, and it is constrained to 2D, retro-style, grid-based games.

There are other models that are more abstract, such as a flow chart of possible choices, actions, and goals that represent a given scenario in a game \cite{paschali2020metric}. While this system constrains the designer to a more narrative style of "gameplay," it is a useful example of a more abstract, graph-based, component-based model of games. Similarly, Game Maker's Toolkit models game economies as a graph of resource gains, losses, and conversions \cite{GMTK}.

\subsection{Generating Games}
Generating game content has been accomplished for various games with various techniques, from random number generation to neural systems \cite{pcgsurveyandgamedesign}, but we focus on game design in this review. Generating new games using genetic techniques (i.e., mutating and evolving designs) and game-playing algorithms with VGDL has also been tried \cite{GeneratingWithVGDL}. This approach led to some interesting designs but relatively poor designs. Our work similarly uses game-playing and genetic algorithms to generate games with our own game description approach. Previous systems design generators/evolvers have been attempted with many constraints (e.g., only for Pacman), also using genetic techniques \cite{pcgsurveyandgamedesign}.

Alternatively to generating a game, Cicero was an AI system using VGDL that assisted human designers with tasks like bug fixing, recommending sprite locations, and recommending interactions \cite{CiceroGameDesignAI}. It does so by searching through the design and finding associations between entities. Like the other works discussed, it is constrained by VGDL, and it also does not have a proper generative system. However, it focuses primarily on the co-creative aspect of these generators and accomplishes great utility that many purely generative systems do not. In our work, we also strive to enable a strong co-creative experience.

\subsection{Evaluating Games}
When generating game systems, especially genetically, it is useful to have a method for the AI agent to evaluate a given game. One approach that serves as a good approximation is the performance of game-playing algorithms as measured by the in-game score as in \cite{GeneratingWithVGDL}, though the authors stated that this approach needed refinement with other metrics. Some metrics of a good game mentioned throughout the literature include "quantity, playability, learnability, difficulty, freshness, interestingness, and utility" \cite{pcgsurveyandgamedesign}. One set of metrics presented for a graph-based, narrative-like model of game scenarios, along with calculations that can be performed on a scenario graph, is re-playability, interactivity, characters' interaction, content, curiosity, desirability to keep playing, and level of narrative \cite{paschali2020metric}.

\section{Methods}
This work expands on the initial Creative Wand framework \cite{Lin_Agarwal_Riedl_2022} through a new creative domain beyond storytelling--namely, game design. Below we present our approach to modeling this domain, creating an AI agent that can work alongside a human designer in this domain, and conducting a study with human subjects.

\subsection{Creative Wand Framework}
\begin{figure*}[h]
    \centering
    \includegraphics[scale=0.4]{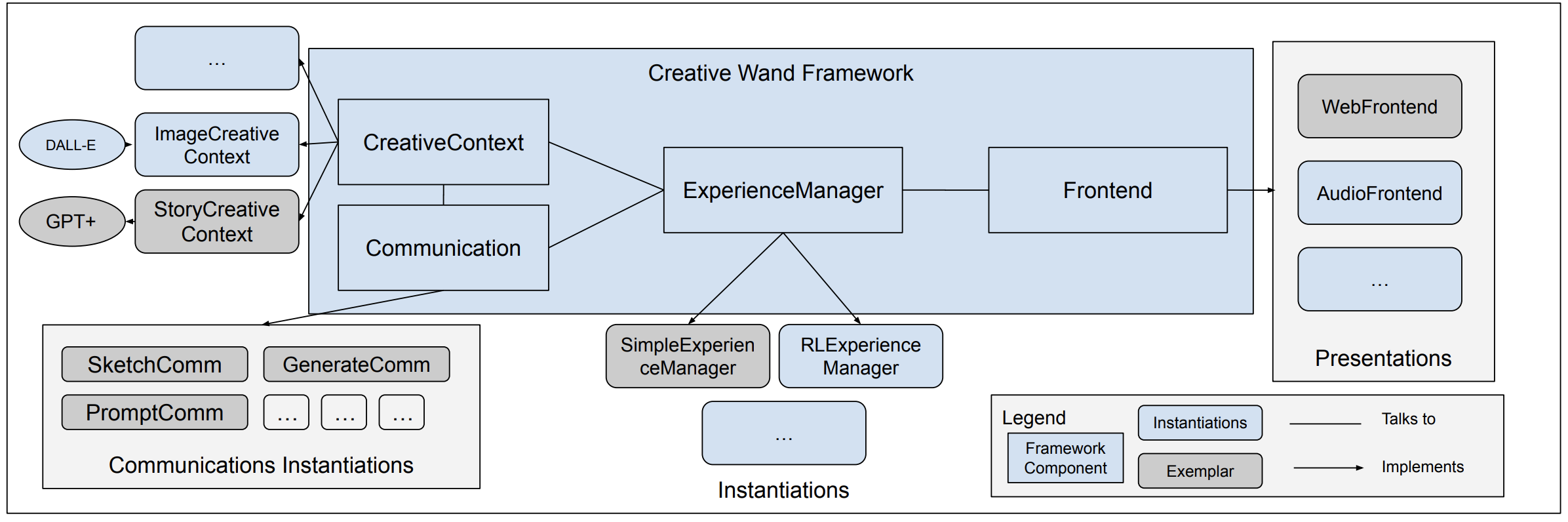}
    \caption{The structure of the Creative Wand framework \cite{Lin_Agarwal_Riedl_2022}. This work uses a creative context with game design abilities, instead of storytelling or art, and communications and a front end to facilitate interacting with that system.}
    \label{fig:CreativeWandFramework}
\end{figure*}

This work makes use of Creative Wand, a framework for building and studying co-creative applications \cite{Lin_Agarwal_Riedl_2022}. The framework consists of several modular components, shown in Fig. \ref{fig:CreativeWandFramework}.

While the original work \cite{Lin_Agarwal_Riedl_2022} focused on storytelling, this work swaps out the creative context for game design and makes changes to the front end as well, which are discussed below. Creative Wand also defines three dimensions of communications between humans and AI in co-creative settings, which informed the abilities given to the novel game design AI agent in this work. We strive to cover these dimensions to allow for a strong co-creative experience, to study our AI system more holistically, and to continue studying co-creative communication itself. These dimensions are:
\begin{itemize}
    \item \textbf{Reflective/Elaborative}: focusing on existing or new content.
    \item \textbf{Local/Global}: focusing on a small part or the whole of the creative work.
    \item \textbf{Human/Agent-Initiated}: whether the human or the AI started the communication \cite{Lin_Agarwal_Riedl_2022}.
\end{itemize}

\subsection{Modeling Games Generically}
One reason to consider game design as the next domain to expand Creative Wand to after storytelling was that, just like storytelling, it can be text-based, making it easier to prototype user interfaces and AI agents and to build off of the original work on Creative Wand \cite{Lin_Agarwal_Riedl_2022}. Additionally, lots of work has been done in procedural content generation (PCG) and game AI, but not as much in AI for game design and game economics. The gaming industry seems like one of the most immediately useful domains for co-creative AI (if mature) due to the lengthy process of game development. In this work, when we refer to "game design," we refer to the early process of thinking about mechanics, systems, and gameplay loops, and not designing individual pieces of content, like levels or narratives. This can also be thought of as system design \cite{pcgsurveyandgamedesign}.

In order to build a co-creative system that can be used to design games, both by human and by AI designers, we must model games generically and abstractly. That is, we must model them in such a way that we can represent a vast range of games (ideally, any kind of game). While there are co-creative storytelling systems that can generate stories about multiple topics \cite{lin2021plug}, most previous work involving game generation only works on a specific type of game at a time, such as Lode Runner \cite{bhaumik2021lode} or Super Mario Bros. \cite{guzdial2016game}. That approach can yield effective results for generating content for a specific kind of game, such as where to place a block so that Mario can jump on top of it, but that is not our task. Instead, we aim to generate, evaluate, and/or balance the systems and mechanics of the game. A human game designer could then take those systems and mechanics and build levels, stories, and the rest of the actual playable game.

One method of modeling games generically is the Video Game Description Language (VGDL) \cite{VGDL}. However, VGDL is relatively unstructured and failed to easily generate playable games \cite{GeneratingWithVGDL}. VGDL is also only suited for 2D, grid-based, retro-style game levels \cite{CiceroGameDesignAI}, so it is not truly general, and once again, it is more focused on level creation than game mechanic/system/loop design itself.

We do not attempt to replace any of these existing approaches, but rather fill a different niche. Our attempt to create such a generic model for game design is inspired by the concept of resource flows as presented in \cite{GMTK}. Though the presented approach is intended for economies in games specifically, mechanics can generally can be thought of economies as well. As an example, a sprinting mechanic transitions a player from a walking state to a running state at the cost of stamina, and while at the walking state, a time resource is converted into the stamina resource. Using these concepts, we define each game as having:
\begin{itemize}
    \item \textbf{Resources}: any consumable the player has, such as money, keys, stamina, time, etc.. The player carries an "inventory" of resources around the game with a maximum capacity for each resource.
    \item \textbf{Actions}: the moves a player can make. Each action can consume some number of some resource.
    \item \textbf{States}: an abstract "game state," to be used freely by the designer. It can represent anything, such as "at the store" or "at the peak of the player's jump." There must be a starting state for the game, and each state has an "importance" value, meaning how important of a goal in the game this state is. This allows flexibility in defining goals.
    \item \textbf{Transitions}: these define how when at one state, which actions can be available, and to which state performing an action takes the player.
    \item \textbf{Taps}: these attach to states. When arriving at the state, the player gains a certain amount of a certain resource.
    \item \textbf{Drains}: these attach to states. When arriving at the state, the player loses a certain amount of a certain resource.
    \item \textbf{Converters}: attach to states. When arriving at the state, the player gains a certain amount of a certain resource and loses a certain amount of another resource.
\end{itemize}

\begin{figure*}[h]
    \centering
    \includegraphics[scale=0.3]{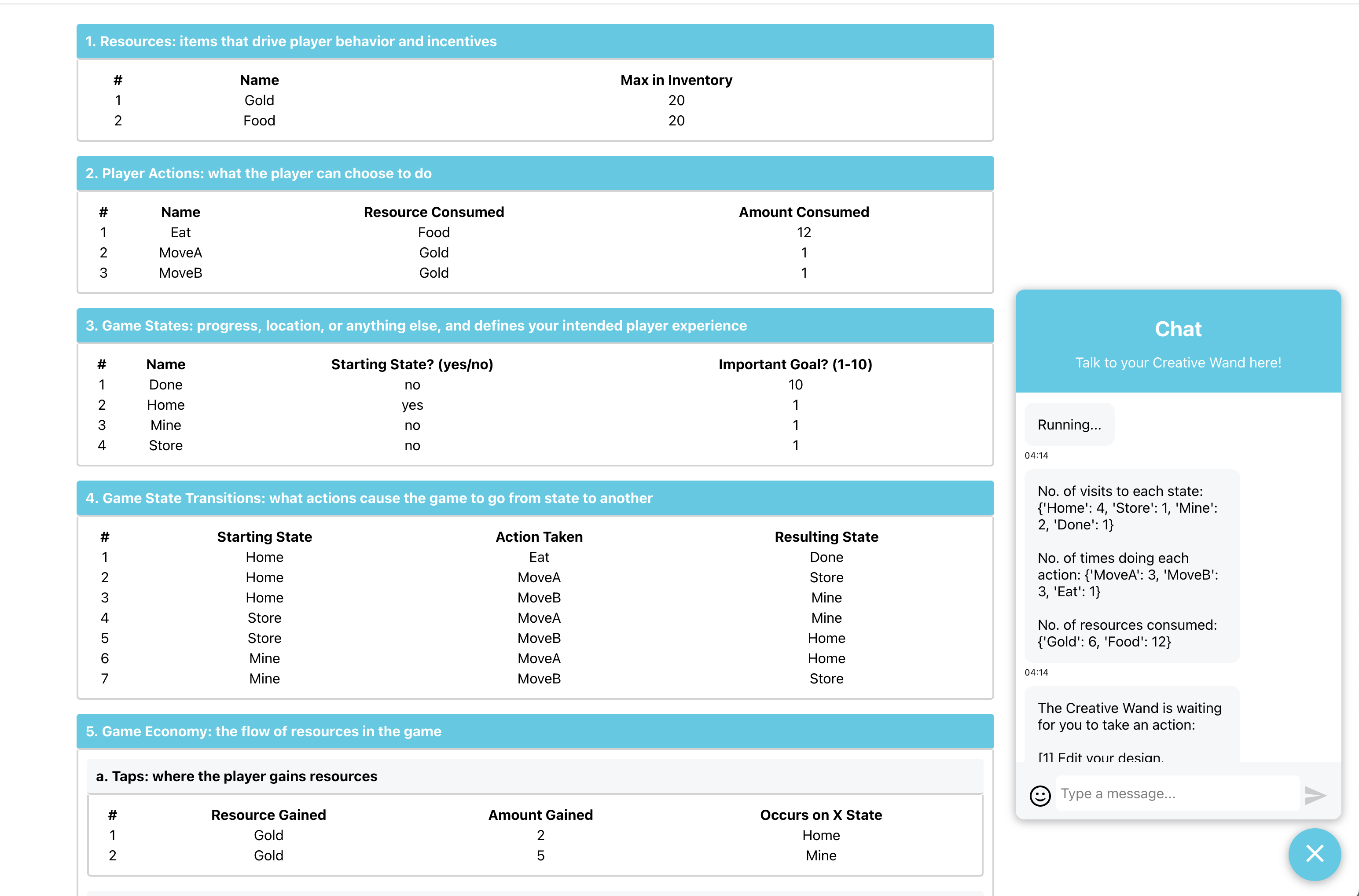}
    \caption{An example of the game design domain in Creative Wand.}
    \label{fig:UIScreenshot}
\end{figure*}

This model, like the others, cannot cover every possible game. For example, we cannot simulate AI characters or in-game processes independent of the player. However, it is sufficient for modeling many individual mechanics and general systems and serves as a simple proof-of-concept for this abstracted kind of model. This approach turns a game mechanic into a specialized state machine with a form of memory. And since states can represent anything at all, it gives the designer flexibility to wire together with transitions all the components they create into a vast range of game designs, with granularity as desired. Most importantly, the model does not constrain the designer to any specific game, style, or genre. An example of what defining these components would look like is shown using the Creative Wand framework in Fig. \ref{fig:UIScreenshot}.

\subsection{Game Domain AI Agent}
After modeling the domain, an AI agent that could work with game designs was required to collaborate with a human designer.

\subsubsection{Game Evaluator}
The first capability implemented for the agent was the ability to evaluate a game design. In the context of the Creative Wand framework, this is a reflective and global communication \cite{Lin_Agarwal_Riedl_2022}. This is because it deals with existing content in the creative work and deals with it as a whole, instead of focusing on certain parts. With a basic implementation, it is human-initiated, but with further development of Creative Wand, it is possible to create an agent-initiated evaluation communication (this applies to all the communications discussed).

To do so, the agent reads the game design components, which can be stored like a JSON object in code. Then, it simulates "playing" the game using a simple Q-learning \cite{watkins1992q} algorithm making use of states, actions, and transitions, with resource-related components constraining the system. The simulated play-through only terminates when the max number of epochs is crossed or there are no valid actions left. This is to allow for open-ended game designs and to see what a player might or could do in a "sandbox" with a certain design. While simulating a play-through of the game, starting at the designated start state, the agent keeps track of a count of each action, state, and resource that the player uses throughout the game. It also keeps track of the path of states taken and the total reward. These statistics are returned to the user to provide insight into how a player might approach the designed game. 

The evaluator uses a set of metrics that approximate what a real game player might care about. These metrics are used to calculate reward at each time step (each time the agent reaches a state) in the Q-learning algorithm. The metrics we propose are:
\begin{itemize}
    \item \textbf{Goal Importance} = designer's "goal importance"
    \item \textbf{State Repetition} = no. of times this state is visited (penalty)
    \item \textbf{State Novelty} = if this state has never been visited so far
    \item \textbf{Action Repetition} = no. of times this action has been used (penalty)
    \item \textbf{Action Novelty} = if this action has never been used so far
    \item \textbf{Resource Gains} = no. of resources gained
    \item \textbf{Resource Losses} = no. of resources lost (penalty)
    \item \textbf{Reward Consistency} = 
sum(reward[-len(rewards) //2 :] 
        
        (rewards having spikes in reward and penalize a consistent reward)
    \item \textbf{Interactivity/Choices Available} = no. of available actions
    \item \textbf{Curiosity/Unvisited States} = no. of unvisited states connected to current state
\end{itemize}

There are other works that evaluate games, but most usually evaluate from a designer's perspective--that is, looking at the whole game--when a player's perspective may be more useful. For example, \cite{paschali2020metric} presents metrics that use calculations on the entire game graph, when in reality, a player would not know everything about the game immediately. Some metrics are still applicable, however, and we take curiosity and interactivity from \cite{paschali2020metric}. The rest we propose, inspired by common ideas and domain experts on player behavior, revolving around goals, resources, novelty/repetition, and reward spacing \cite{GMTKengagement}.

Of course, a Q-learning agent may very well not play like a human would, and since this is a co-creative system, we add a level of user control to this function. Users can specify a weight, or multiplier, for each of the listed metrics, controlling what the agent considers when evaluating the game (and in a way, being able to "play test" with players who value different things). In addition to referencing domain experts, we also generate 10000 random game designs with the generator described below and calculate the expressive range of our metrics to gain insight into how much information about games these metrics can capture.

\subsubsection{Genetic Game Balancer}
The next capability implemented was for the AI agent to balance the numerical values in the design without modifying which components made up the design. This includes resource maximums, action costs, and amounts gained or lost at taps, drains, and converters. This ability allows for a reflective and local communication because it focuses on tweaking existing content and only very specific parts of it \cite{Lin_Agarwal_Riedl_2022}. It is accomplished by using a genetic algorithm \cite{whitley1994genetic} that creates random mutations in these numbers within the original design and uses the discussed evaluation method to select population members to move forward. At the end of its iterations, it returns the design with highest total score. However, a chromosome is a set of lists of numbers (one for each component category), rather than 0s and 1s in order to maintain a valid structure and store specific values. Again, as this is a co-creative system, we introduce two user controls. First, this communication offers the ability to set weights for the metrics used in evaluation as discussed before. Second, it offers the option to ask the user to select their favorite population member, instead of using the evaluation function, every several iterations.

\subsubsection{Genetic Game Generator}
The final capability implemented is for the AI agent to modify the components in the design and even add more (essentially, a constrained game design generator). With a control allowing users to select which component types can be modified, this communication can be both local and global. Since it is adding new content, it is elaborative. Once again, it can be either human- or agent-initiated, depending on Creative Wand itself, but for this study, it is human-initiated like the other communications \cite{Lin_Agarwal_Riedl_2022}. Like the genetic game balancer, this is accomplished using a genetic algorithm starting with the provided design and offers similar controls with metric weights and human intervention, returning the highest scoring design in the end. Chromosomes here are represented as a list of string attributes (one for each component type) and mutations are changing or adding new pieces in constrained yet random ways to preserve a valid structure. Additionally, the chromosomes in the genetic algorithm are variable length in order to allow addition of new components.


\subsection{Study}

\subsubsection{Expressive Range}
First, we conducted an expressive range analysis of the generation system. 100,000 random designs were generated (with some cap on the number of each component to keep the analysis in scope), and with roughly 33,000 selected as playable, the mean score throughout the simulated playthrough for each game was collected. This allowed for studying the distributions of each metric and the ability of the generator to express a wide variety of designs.

\subsubsection{Controllability with Metric Weights}
 The system offers several controls already, such as choosing to balance or generate and choosing to freeze certain categories, and they work by the fact that they are implemented. However, for setting metric weights, we need to ensure that the system responds properly to changes. If the user asks for a design weighing a certain metric more or less, can the system give the user what they want, or will it just output anything? Expressive range shows that different games have different metric scores; controllability shows that different metric scores give different games.

With the same process as the expressive range analysis, 25 random games were generated and the metric scores stored. For each game, we set a given metric to a given weight while all others remain set to 1, run the balancer, and measure the change in the metric score. We repeat this with all metrics and with weights -100 and 100. We also evaluate the generator with this same methodology. To analyze controllability of each metric, we consider the score difference between the two weights for each generated game. The reason for this methodology is to eliminate the variability caused by different game designs; balancing one design may cause a metric score to go down no matter what, but what matters is the degree to which that score change differs depending on the set weight.

\section{Results}

\subsection{Expressive Range Analysis}

\begin{figure*}[h]
    \includegraphics[width=.24\textwidth]{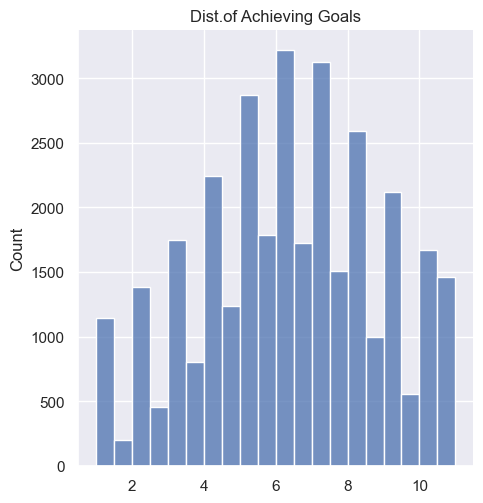}\hfill
    \includegraphics[width=.24\textwidth]{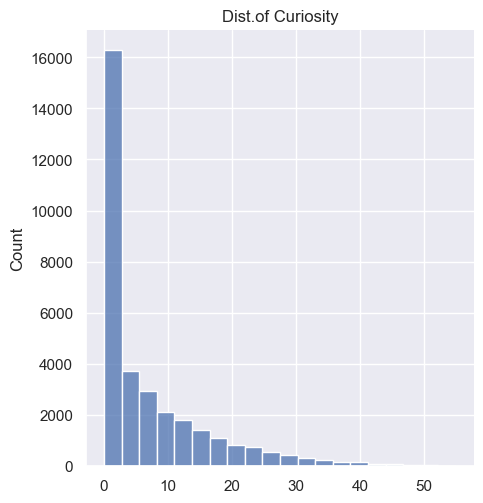}\hfill
    \includegraphics[width=.24\textwidth]{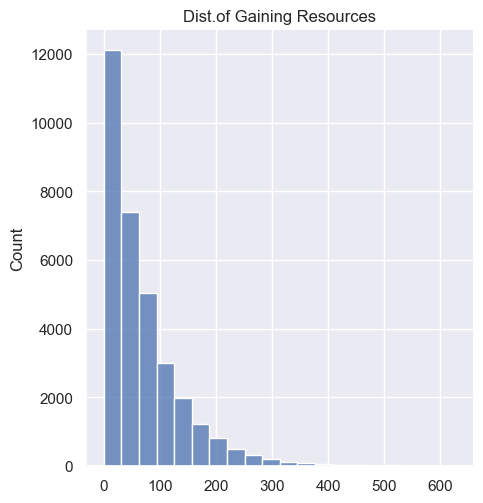}\hfill
     \includegraphics[width=.24\textwidth]{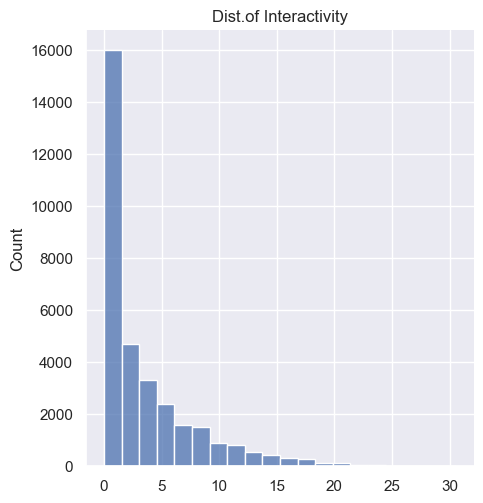}
     \\[\smallskipamount]
    \includegraphics[width=.24\textwidth]{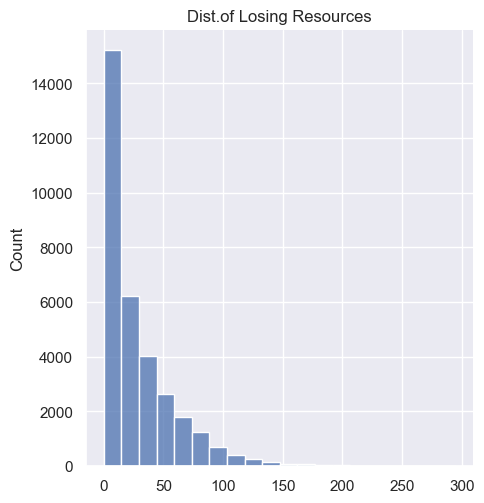}\hfill
    \includegraphics[width=.24\textwidth]{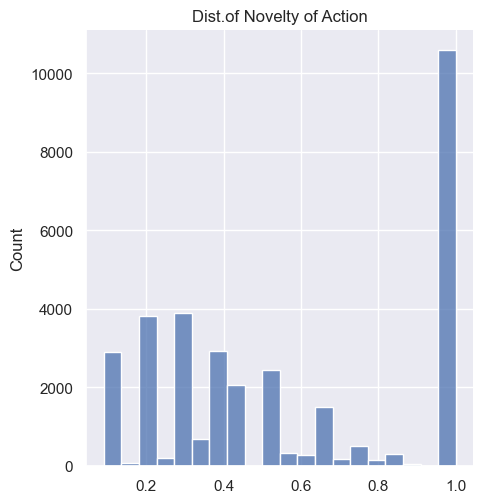}\hfill
    \includegraphics[width=.24\textwidth]{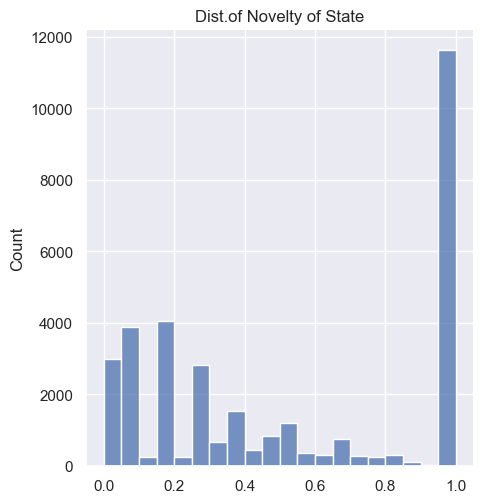}\hfill
    \includegraphics[width=.24\textwidth]{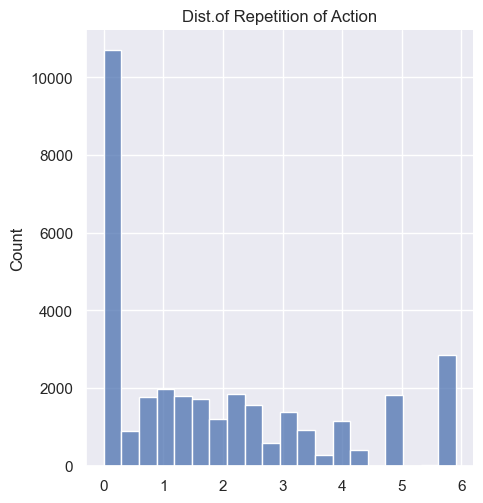}
    \\[\smallskipamount]
    \includegraphics[width=.24\textwidth]{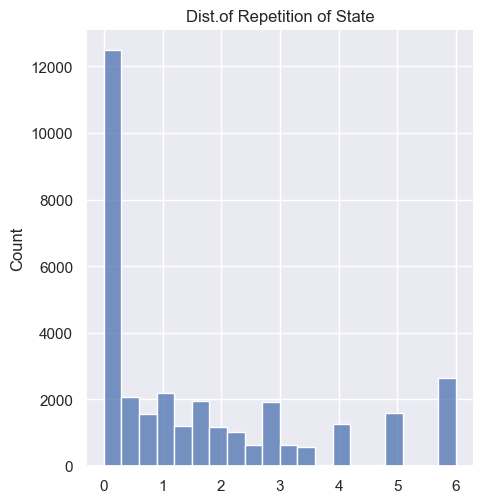}
    \includegraphics[width=.24\textwidth]{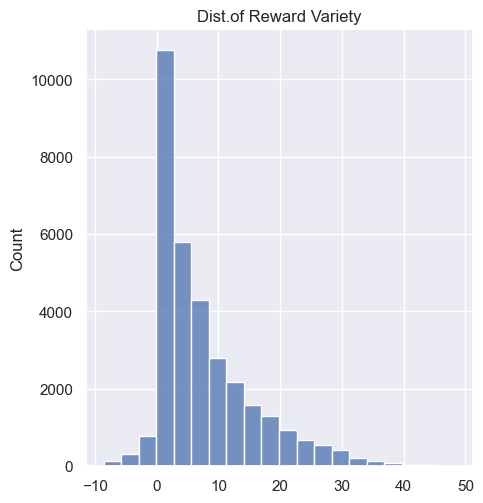}
    \caption{Distributions of mean scores for all 10 metrics across $\sim$33k generated game designs. They all follow one of normal, uniform, or Pareto distributions, often with the addition of a spike at 0 or 1.}
    \label{fig:ExpressiveRangeDists}
\end{figure*}

\begin{table*}[!ht]
    \centering
    \begin{tabular}{|p{20mm}|l|l|l|l|p{20mm}|p{70mm}|}
    \hline
        \textbf{Metric} & \textbf{Min} & \textbf{Mean} & \textbf{Max} & \textbf{Variance} & \textbf{Distribution Description} & \textbf{Reasoning} \\ \hline
        Goal Importance & 1 & 6.2 & 11 & 6.4 & Normal & Adding together states w/ "random" values assigned by designer \\ \hline
        State Repetition (-) & 0 & 1.7 & 6 & 3.8 & Uniform w/ spike at 0 & Which state is visited and no. of times each state is visited is random, but every state is new to start \\ \hline
        State Novelty & 0 & 0.5 & 1 & 0.2 & Normal w/ spike at 1 & More likely to have a mix of new and visited states throughout a game, but every state is new to start \\ \hline
        Action Repetition (-) & 0 & 1.9 & 5.9 & 3.8 & Uniform w/ spike at 0 & Which action is used and no. of times each action is used is random, but every action is new to start \\ \hline
        Action Novelty & 0.1 & 0.6 & 1 & 0.1 & Normal w/ spike at 1 & More likely to have a mix of new and unused actions throughout a game, but every action is new to start \\ \hline
        Resource Gains & 0 & 68.1 & 596.5 & 4865.3 & Pareto & Like real life, only a a few sets of compounding causes lead to the most resource gains \\ \hline
        Resource Losses (-) & 0 & 26.9 & 306.5 & 975.1 & Pareto & Like real life, only a a few sets of compounding causes lead to the most resource losses \\ \hline
        Reward Consistency (-) & -8.5 & 7.3 & 49.6 & 67.3 & Pareto in both +/- directions & Like real life, only a a few sets of compounding causes lead to the most resource gains/losses, so only a few have extremely consistent reward variety \\ \hline
        Interactivity & 0 & 3.3 & 29 & 18.6 & Pareto & A lot of factors must compound to have a lot of choices available throughout the game \\ \hline
        Curiosity & 0 & 6.9 & 55.5 & 85.2 & Pareto & A lot of factors must compound to have a lot of new states available throughout the game \\ \hline
    \end{tabular}
    \caption{Statistics and distribution descriptions from the distributions of mean scores for all 10 metrics across $\sim$33k generated game designs. (-) indicates that this metric is treated as a penalty instead of a reward.}
    \label{table:ExpressiveRangeStats}
\end{table*}

In Fig. \ref{fig:ExpressiveRangeDists}, we graph the distributions for each metric across the generated games. In Table \ref{table:ExpressiveRangeStats}, we present some corresponding statistics (mean, variance, maximum, and minimum), describe each distribution, and provide some reasoning for why the distribution takes that shape. The purpose of this analysis is simply to observe whether the generator outputs designs in an reasonable way across metrics. This is important for the sake of human-controllability when assigning metric weights (we can use this information to normalize raw scores and make sure applying a weight is intuitive) and for demonstrating that metric scores are meaningful and not random, unchanging, or useless in some other way. The exact distribution does not matter as much.

Notably, all the metrics approximately follow some form of either normal, uniform, or Pareto distributions. Some metrics have spikes at 0 or 1 due to the nature of the game playing environment; for example, every state in a game is guaranteed to be novel at first to the game playing agent, so state repetition has a spike of 0 values. All of these distributions seem reasonable for their corresponding metric; for example, resource gains follow a Pareto distribution, just as wealth accumulation does in real life. Overall, the generator is capable of creating a broad range of games with well-distributed levels of each metric. Likewise, it is clear that the chosen 10 metrics vary predictably across a wide range of designs. Knowing that the generator and its metrics can generate games reasonably well, we can next evaluate its controllability.

\subsection{Controllability with Metric Weights}

\begin{figure*}[h]
    \includegraphics[width=.5\textwidth]{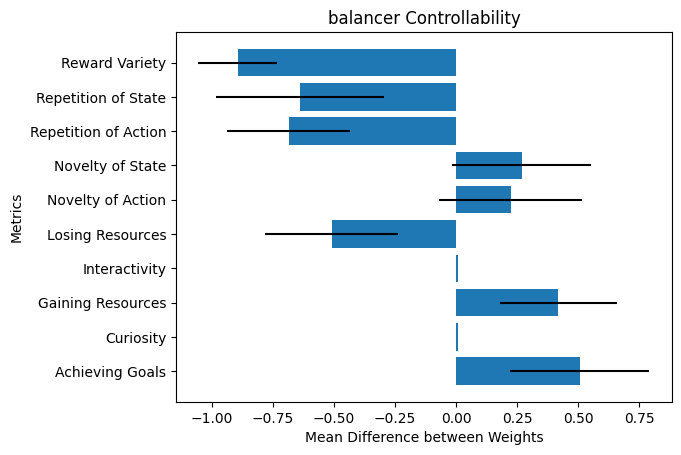}\hfill
    \includegraphics[width=.5\textwidth]{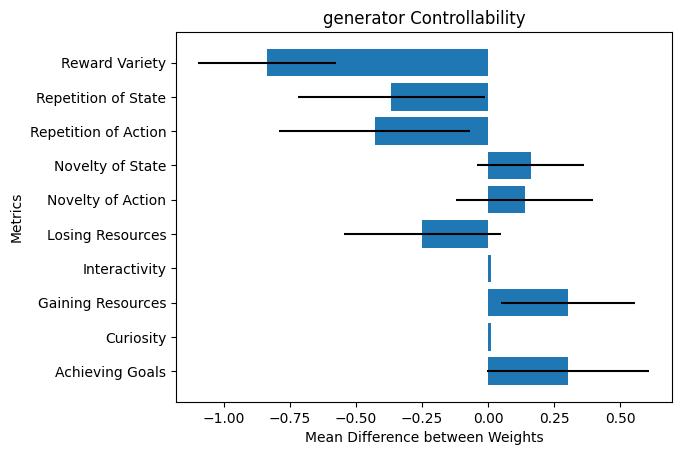}\hfill

    \caption{Metric controllability for the balancer and generator, represented by the mean difference in scores between the two weights tried. Error bars represent standard deviation. Different metrics are shifted positively or negatively as expected since some are considered rewards and others penalties.}
    \label{fig:ControllabilityPlots}
\end{figure*}

\begin{table}[h]
    \centering
    \begin{tabular}{|l|l|l|}\hline
    \textbf{Metric}      & \textbf{Balancer} & \textbf{Generator} \\\hline
    Reward Variety       & 2.275e-11**       & 8.727e-9**         \\\hline
    Repetition of State  & 5.548e-6**        & 0.002**            \\\hline
    Repetition of Action & 1.852e-7**        & 0.0006**           \\\hline
    Novelty of State     & 0.011*           & 0.052*             \\\hline
    Novelty of Action    & 0.029*           & 0.106              \\\hline
    Losing Resources     & 2.202e-5**        & 0.012*            \\\hline
    Interactivity        & 0.5               & 0.5                \\\hline
    Gaining Resources    & 0.0001**          & 0.003**            \\\hline
    Curiosity            & 0.5               & 0.5                \\\hline
    Achieving Goals      & 5.013e-5**        & 0.007**           \\\hline
    \end{tabular}
    \caption{p-values of metric controllability. * = $p < 0.1$, ** = $p < 0.01$}
    \label{tab:ControllabilityPVals}
\end{table}

In Fig. \ref{fig:ControllabilityPlots}, a visualization of the controllability of each metric is presented, along with significance statistics in Table \ref{tab:ControllabilityPVals}. First, it is clear that the interactivity and curiosity metrics, adapted from \cite{paschali2020metric}, are not controllable by either the balancer or the generator. Perhaps this is due to these metrics originally being designed for analyzing a game as a whole, rather than at individual, localized steps of a playthrough. In general, all other metrics appear controllable by both. A few, such as novelty of action and novelty of state have higher $p$-values, but perhaps a more extreme weight would lead to a larger difference. These metrics might be slightly less controllable due to relying on generating content and the evaluation agent being able to access that new content--a task that would benefit from the balancer and generator working in conjunction. In general, a system that can simultaneously balance numbers and generate new components may lead to even higher controllability for all metrics (and may cover the other small differences in controllability in some metrics between the generator and balancer, as shown in Fig. \ref{fig:ControllabilityPlots}). All other metrics are significantly controllable ($p < 0.05$) in both the balancer and generator, and scores shift in the expected direction depending on whether the metric is a penalty or reward. Overall, human designers can significantly and expectedly control the output of our AI system by setting weights for different metrics used in game design evaluation.

\section{Discussion}

Through expressive range analysis, we found our generator able to express a predictably-distributed range of games as described by our proposed metrics. Through analysis of the controllability of each metric in the balancer and generator, we found that most of the metrics are controllable, but a few are not. These non-controllable metrics still may be useful for the evaluator (and by extension, the balancer and generator), and even as a point of feedback for the human designer. While the system would also work fine without these specific metrics as it has plenty of others, future work should explore new or modified metrics to refine the system for co-creative applications as controllability is crucial. Additionally, combining the balancing and generation systems into one may lead to higher controllability. The amount by which a weight changes a metric score is variable; further work is needed to normalize and scale that for users.

In this study, we evaluated our novel game design generator through automated means, answering the questions of how expressive and how controllable it is. In future studies, we intend to answer how "good" the system is by evaluating how human subjects work with the system, similar to the original Creative Wand study \cite{Lin_Agarwal_Riedl_2022}. Such a study could focus on the outputs of the system: how human's perceive the evaluations provided, if humans agree that the balancer makes a design more balanced, and if humans prefer designs that the generator adds to. It is also important to study the effect of different communications (as with Creative Wand \cite{Lin_Agarwal_Riedl_2022}) with the system on the creative process. A human subject study will also help evaluate our proposed model of game design itself by providing insight into how humans interact with and use the model.

The experiments conducted in this study were also limited by constraints on the size of designs, the number of designs tested, and the number of iterations in our algorithms due to time and computational constraints. Additionally, the current iteration of our game design model could be both expanded and simplified, such as by combining all resource flow components and by adding events, conditions, timing, initial inventory, and other useful features for game system/mechanic design. Our model and metrics can even be separated out and used with other forms of AI agents, such as large language models, in future work.

\section{Conclusion}

This study shows that an abstract, component-based model of game system designs, based on a state machine and resource flows, can express a wide range of games as captured by a set of metrics, evaluated through a player simulation. These metrics are human-controllable through weighting and guide the output of a genetic game balancer and generator. Our proposed model, metrics, and AI agent system are a successful first approach, and a strong starting point for further development and experimentation, to co-creative game design.

\bibliographystyle{plain} \bibliography{refs}

\begin{thebibliography}{10}

\bibitem{bhaumik2021lode}
Debosmita Bhaumik, Ahmed Khalifa, and Julian Togelius.
\newblock Lode encoder: Ai-constrained co-creativity.
\newblock In {\em 2021 IEEE Conference on Games (CoG)}, pages 01--08. IEEE,
  2021.

\bibitem{guzdial2016game}
Matthew Guzdial and Mark Riedl.
\newblock Game level generation from gameplay videos.
\newblock In {\em Twelfth artificial intelligence and interactive digital
  entertainment conference}, 2016.

\bibitem{pcgsurveyandgamedesign}
Mark Hendrikx, Sebastiaan Meijer, Joeri Van Der~Velden, and Alexandru Iosup.
\newblock Procedural content generation for games: A survey.
\newblock {\em ACM Trans. Multimedia Comput. Commun. Appl.}, 9(1), feb 2013.

\bibitem{Lin_Agarwal_Riedl_2022}
Zhiyu Lin, Rohan Agarwal, and Mark Riedl.
\newblock Creative wand: A system to study effects of communications in
  co-creative settings.
\newblock {\em Proceedings of the AAAI Conference on Artificial Intelligence
  and Interactive Digital Entertainment}, 18(1):45--52, Oct. 2022.

\bibitem{lin2021plug}
Zhiyu Lin and Mark Riedl.
\newblock Plug-and-blend: A framework for controllable story generation with
  blended control codes.
\newblock {\em arXiv preprint arXiv:2104.04039}, 2021.

\bibitem{CiceroGameDesignAI}
Tiago L. d.~A. Machado.
\newblock {\em Cicero - An AI-Assisted Game Design System}.
\newblock PhD thesis, 2019.
\newblock Copyright - Database copyright ProQuest LLC; ProQuest does not claim
  copyright in the individual underlying works; Last updated - 2022-12-20.

\bibitem{GeneratingWithVGDL}
Thorbjørn~S. Nielsen, Gabriella A.~B. Barros, Julian Togelius, and Mark~J.
  Nelson.
\newblock Towards generating arcade game rules with vgdl.
\newblock In {\em 2015 IEEE Conference on Computational Intelligence and Games
  (CIG)}, pages 185--192, 2015.

\bibitem{paschali2020metric}
Eleni Paschali, Apostolos Ampatzoglou, Remi Escourrou, Alexander
  Chatzigeorgiou, and Ioannis Stamelos.
\newblock A metric suite for evaluating interactive scenarios in video games:
  an empirical validation.
\newblock In {\em Proceedings of the 35th Annual ACM Symposium on Applied
  Computing}, pages 1614--1623, 2020.

\bibitem{COFI}
Jeba Rezwana and Mary~Lou Maher.
\newblock Designing creative ai partners with cofi: A framework for modeling
  interaction in human-ai co-creative systems.
\newblock {\em ACM Trans. Comput.-Hum. Interact.}, feb 2022.
\newblock Just Accepted.

\bibitem{GMTKengagement}
Game~Maker\textquotesingle s~Toolkit.
\newblock How to keep players engaged (without being evil), 2018.
\newblock {https://www.youtube.com/watch?v=hbzGO\_Qonu0\&t=68s}.

\bibitem{GMTK}
Game~Maker\textquotesingle s~Toolkit.
\newblock How video game economies are designed, 2022.
\newblock {https://www.youtube.com/watch?v=Zrf1cou\_yVo}.

\bibitem{VGDL}
Tommy Thompson, Marc Ebner, Tom Schaul, John Levine, Simon Lucas, and Julian
  Togelius.
\newblock Towards a video game description language.
\newblock {\em Dagstuhl Follow-ups}, 6:85, 11 2013.

\bibitem{watkins1992q}
Christopher~JCH Watkins and Peter Dayan.
\newblock Q-learning.
\newblock {\em Machine learning}, 8(3):279--292, 1992.

\bibitem{whitley1994genetic}
Darrell Whitley.
\newblock A genetic algorithm tutorial.
\newblock {\em Statistics and computing}, 4(2):65--85, 1994.

\bibitem{Yannakakis2014MixedinitiativeC}
Georgios~N. Yannakakis, Antonios Liapis, and Constantine Alexopoulos.
\newblock Mixed-initiative co-creativity.
\newblock In {\em International Conference on Foundations of Digital Games},
  2014.

\bibitem{cocreativedrawing}
Chao Zhang, Cheng Yao, Jianhui Liu, Zili Zhou, Weilin Zhang, Lijuan Liu,
  Fangtian Ying, Yijun Zhao, and Guanyun Wang.
\newblock Storydrawer: A co-creative agent supporting children's storytelling
  through collaborative drawing.
\newblock In {\em Extended Abstracts of the 2021 CHI Conference on Human
  Factors in Computing Systems}, CHI EA '21, New York, NY, USA, 2021.
  Association for Computing Machinery.

\end{thebibliography}

\end{document}